\documentclass[10pt,twocolumn,letterpaper]{article}

\usepackage{cvpr}
\usepackage{times}
\usepackage{epsfig}
\usepackage{graphicx}
\usepackage{subfigure}
\usepackage{caption}
\usepackage{amsmath}
\usepackage{amssymb}
\usepackage{enumitem}

\usepackage{hyperref}
\hypersetup{pagebackref=true,breaklinks=true,letterpaper=true,colorlinks,bookmarks=false}

\cvprfinalcopy 

\def\cvprPaperID{27} 

\ifcvprfinal\fi
\begin{document}

\title{CalliGAN: Style and Structure-aware Chinese Calligraphy Character Generator}

\author{Shan-Jean Wu, Chih-Yuan Yang and Jane Yung-jen Hsu \\
Computer Science and Information Engineering\\
National Taiwan University\\
{\tt\small \{r06944023, yangchihyuan, yjhsu\}@csie.ntu.edu.tw}
}


\maketitle

\begin{abstract}
   Chinese calligraphy is the writing of Chinese characters as an art form performed with brushes so Chinese characters are rich of shapes and details. Recent studies show that Chinese characters can be generated through image-to-image translation for multiple styles using a single model. We propose a novel method of this approach by incorporating Chinese characters' component information into its model. We also propose an improved network to convert characters to their embedding space. Experiments show that the proposed method generates high-quality Chinese calligraphy characters over state-of-the-art methods measured through numerical evaluations and human subject studies.
\end{abstract}

\section{Introduction}
Chinese characters are logograms developed for the writing of Chinese. Unlike an alphabet, every Chinese character has its own meaning and an entire sound. Chinese characters were invented several thousand years ago, initially as scripts inscribed on animal bones or turtle plastrons. Around 300 BC, ink brushes were invented. During the Qin Dynasty (221 BC to 206 BC), Chinese characters were first standardized as the Qin script. Thereafter, they were developed into different forms in the long history such as the clerical, regular, semi-cursive, and cursive scripts.

\begin{figure}[t]
\centering
\includegraphics[width=\linewidth]{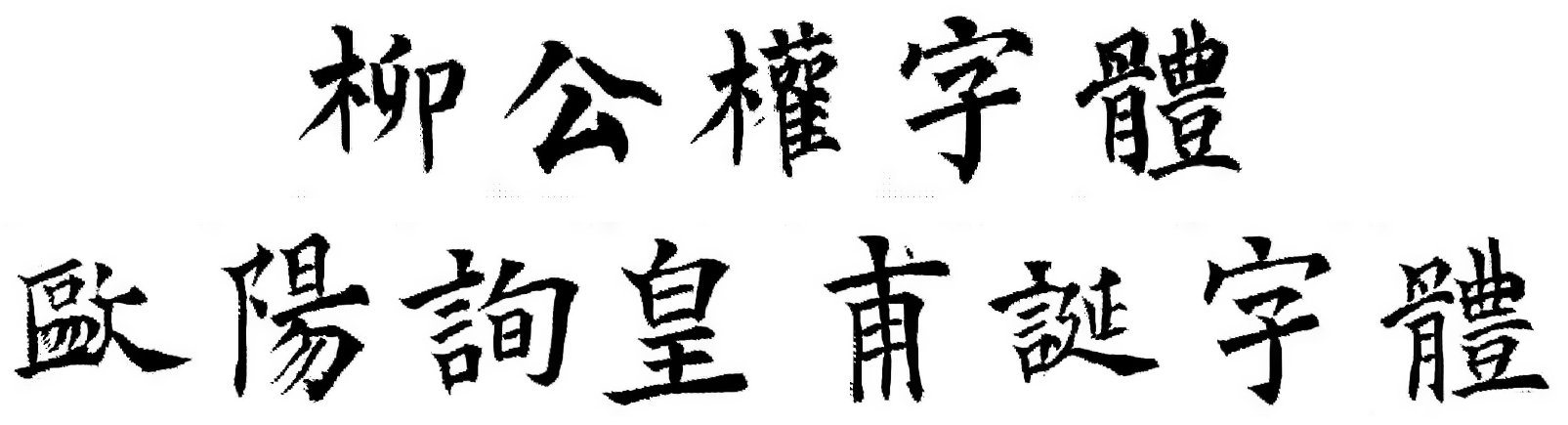}
\caption{Results generated by the proposed method. The style used to generate characters of the upper row is style 2 (Liu Gongquan) and of the lower row is style 3 (Ouyang Xun–Huangfu Dan Stele).}
\label{fig:intro}
\end{figure}

Along with its long history, Chinese calligraphy belongs to the quintessence of Chinese culture. While calligraphers use brushes to write characters, they also embody their artistic expressions in their creatures. Therefore, every brush-written character image is unique and irregular like a picture. In contract, fonts are created by companies and font-rendered images often contain common regions such as radicals.
In addition, different fonts cover different numbers of characters. For example, the widely used Chinese font Sim Sun version 5.16 covers 28762 Unicode characters and its extension package version 0.90 covers 42809 rarely used Chinese characters\footnote{\url{https://zh.wikipedia.org/wiki/\%E4\%B8\%AD\%E6\%98\%93\%E5\%AE\%8B\%E4\%BD\%93}}, but some fonts only cover limited numbers of characters.
Brush-written characters, in particular masterpieces, have another problem that some characters become unclear or damaged if their papers and steles decay. The absence of many characters restrains calligraphy beginners from emulating masterpieces and designers from using masters' works. Therefore, there is a need to generate character images like Figure~\ref{fig:intro} and many methods have been published to address this problem.

Because Chinese characters are highly structured, some early developed methods use the split-and-merge approach to decompose a character into strokes, and then assemble each stroke's synthesized calligraphy counterpart into a calligraphy character~\cite{xu2005automatic,xu2009automatic}. However, the approach has a limitation that Chinese characters with complex structures are difficult to be decomposed automatically, and require manual decomposition for certain styles such as the cursive script~\cite{xu2007intelligent}. 

With the advance of neural networks and computer vision, a study called style transfer, which aims to add painters' artistic styles to photos captured by cameras, shows remarkable success~\cite{gatys2016image,johnson2016perceptual,ulyanov2016texture,dumoulin2016learned, huang2017arbitrary}. Style transfer gradually evolves to image-to-image translation~\cite{pix2pix,zhu2017unpaired,yi2017dualgan,kim2017learning,liu2017unsupervised,dtn,stargan}, which aims to not only add style details to target images but also convert objects from one domain to another, for example, horses to zebras, and vice versa. 
Because every Chinese calligrapher has his or her own style in terms to form strokes and shapes, generating calligraphy characters can be viewed as translating character images from one domain to another ~\cite{chang2018chinese,chang2018generating,wen2019handwritten,jiang2017dcfont,jiang2019scfont,sun2018pyramid,zhang2018separating,sun2017learning}.

A Chinese font can easily render numerous character images. Given two fonts, we can easily obtain numerous well-aligned character pairs. Therefore, it is a practical approach to generate characters by training an image-to-image translation model which take font-rendered character images as input, and  this approach generates the state-of-the-art quality~\cite{chang2018chinese}. 
Compared with font-rendered character images, brush-written character images are more irregular. In addition, they take time and effort to create. To the best of our knowledge, there is no well-defined dataset of brush-written calligraphy character images available, and there is only one existing paper using brush-written calligraphy character images to conduct experiments~\cite{aegg}. This paper is the second to deal with this image type.

Using brush-written images, we develop a method of multi-style image-to-image translation. We define styles basically as calligraphers' identities. If a calligrapher has distinct styles at different periods of creation, we define multiple style labels for that calligrapher. To validate the developed method, we conduct head-to-head comparisons with an existing method. 
To sum up, this paper has two contributions:
\begin{itemize}
    \item While existing multi-font Chinese character generating methods are designed to generate highly different fonts, this paper is the first one dealing styles at the fine-grained level. In addition, this paper is the second paper reporting experimental results of brush-written calligraphy images. Our code and dataset are publicly available to help researchers reproduce our results.
    \item The proposed method has a novel component encoder. To the best of our knowledge, the proposed method is the first to decompose Chinese characters into components and encoder them through a recurrent neural network. The proposed method generates promising images which lead to favorable numerical evaluations and subjective opinions.
\end{itemize}{}

\section{Related Work}
There are numerous methods in the literature generating Chinese character images. The proposed method formulates Chinese character generation as an image-to-image translation problem, and we discuss its related work as follows.

\begin{figure*}[t]
\includegraphics[width=1.0\linewidth]{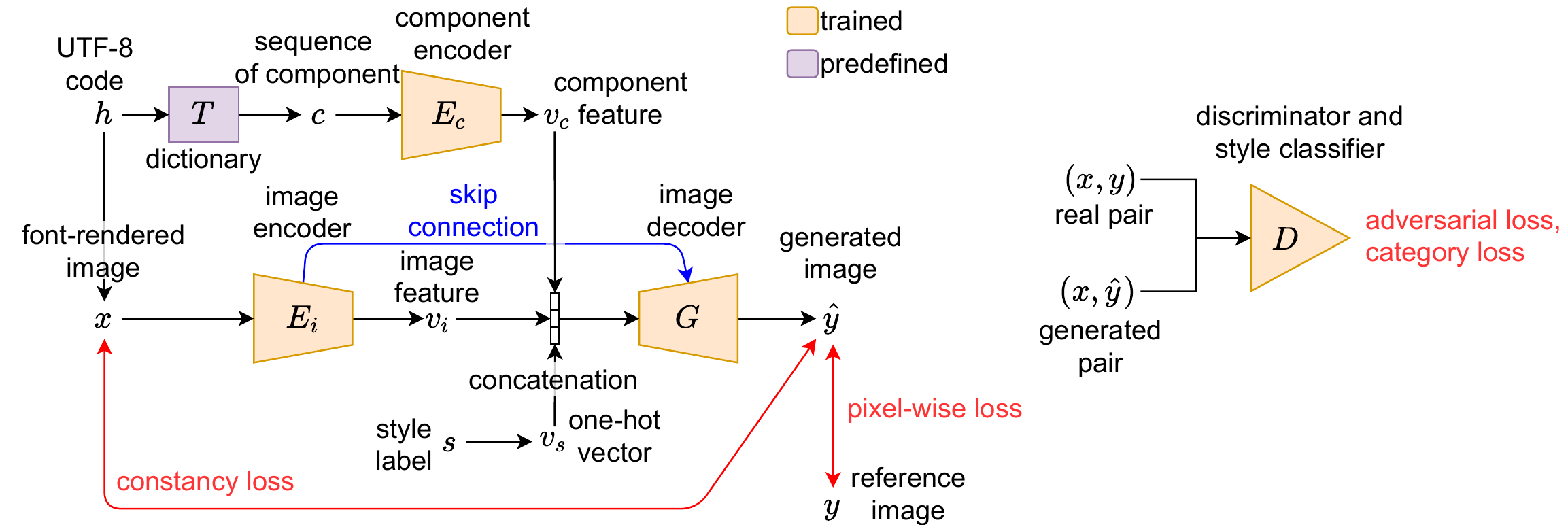}
\caption{Architecture and losses. The proposed CalliGAN is an encoder-decoder-based image translation network with two supporting branches to control styles and structures. CalliGAN has 4 image-based losses: adversarial (Eq.~2), pixel-wise (Eq.~3), constancy (Eq.~4) and category (Eq.~5).}
\label{fig:model}
\end{figure*}

\textbf{\flushleft Image-to-image translation.}
Image-to-image translation is a type of vision and graphics problems. It aims to learn a mapping function between an input image and an output image. There is a wide range of application using this technique such as style transfer, object replacement, season transfer, or photo enhancement.

Numerous image-to-image translation methods have been published, and most of them GAN-based, conditioned on images~\cite{pix2pix,zhu2017unpaired,stargan,starganv2}. Pix2pix~\cite{pix2pix} is the first method capable of doing image-to-image translation. 
Different from preceding neural-network-based style transfer methods, it extracts style representation from a set of target images, which helps pix2pix generate more robust output images than style transfer methods. In addition, its GAN-based adversarial loss prevents its output images from being blurry, and its image quality outperforms most encoder-decoder-based methods~\cite{hinton2006reducing}. 

Pix2pix uses U-Net~\cite{unet} as its generator, which consists an image encoder and a decoder. Between them there is skip connection to preserve visual information through all layers. Pix2pix uses a $l_1$-norm pixel-wise loss to reduce the differences between output and training images.

One of pix2pix's limitations is that it required paired images to train its model. Those pairs are easily available for some applications such as photo-to-sketch transfer, but hard to obtained for other applications such as object replacement. CycleGAN~\cite{zhu2017unpaired} is proposed to overcome the problem by developing two GANs in a cycle. One GAN's output is another GAN's input, and vice versa. The two GANs learns image distributions simultaneously, so that they can use two sets of training images instead of a single set of paired images.

However, CycleGAN can only handle one target domain. To generate images of multiple domains, multiple models are required to be trained individually. StarGAN~\cite{stargan} is proposed to address this issue. It introduces an auxiliary domain classifier and a classification loss to achieve multi-domain translation in a single model. 
The proposed CalliGAN's image generator is similar to pix2pix, and CalliGAN can handle multi-class image-to-image translation like StarGAN.

\textbf{\flushleft Chinese character generation.}
Zi2zi~\cite{zi2zi} is the first method generating Chinese characters using GANs. It translates character images of a source font to multiple target fonts. Based on pix2pix, zi2zi adapts AC-GAN's~\cite{acgan} auxiliary classifier to enable multiple styles generation, and DTN's~\cite{dtn} constancy loss to improve output quality. Zi2zi's output font is controlled by a class parameter formed as a one-hot vector and converted to a latent vector through embedding.

Zi2zi is an open source project, but never published as a paper or technical report. The first paper using GANs to generate Chinese calligraphy characters is AEGG~\cite{aegg}, which is also based pix2pix, but adds an additional encoder-decoder network to provide supervision information in the training process. Unlike zi2zi which can generate multi-class images, AEGG only supports single-class character generation. 

Both of DCFont~\cite{jiang2017dcfont} and PEGAN~\cite{sun2018pyramid} are modified from zi2zi to generate the whole 6763 Chinese characters used in the GB2312 font library from hundreds of training samples.
While PEGAN improves zi2zi by introducing a multi-scale image pyramid to pass information through refinement connections, DCFont incorporates a style classifier pre-trained on 100 fonts to get better style representation. SCFont~\cite{jiang2019scfont} further improves DCFont by adapting a stroke extraction algorithm~\cite{Lian2016} to maintain stroke structures from input to output images.

In contrast to learning translation models between given fonts, both of EMD~\cite{zhang2018separating} and SA-VAE~\cite{sun2017learning} separate content and styles as two irrelevant domains and uses two independent encoders to model them. However, their technical details are different. EMD mixes style and content latent features in a bilinear mixer network to generate output images through an image decoder. Therefore, its training samples are very special. One sample consists of two sets of training images, one for content and another for style.
In contrast, SA-VAE adapts a sequential approach. It first recognizes characters from given images, and then encodes the recognized characters into special codes, which represent 12 high-frequency Chinese characters' structure configurations and 101 high-frequency radicals. SA-VAE shows that domain knowledge of Chinese characters helps improve output image quality.

The proposed CalliGAN shares two common points with existing methods. First, CalliGAN is a GAN-based method, like zi2zi, AEGG, DCFont, and PEGAN. Second, CalliGAN exploits prior knowledge of Chinese characters' structures, like SA-VAE. A clear difference between CalliGAN and SA-VAE is the ways of exploit Chinese characters' structures. SA-VAE only uses characters' configurations and radicals, which are high-level structure information, but CalliGAN fully decomposes characters into components, which offer low-level structure information including the order of strokes. In short, CalliGAN integrates the advantages of GANs which generate realistic images and SA-VAE which preserves character structures.


\section{Proposed Method}
A Chinese character can be expressed in multiple styles, depending on the fonts used to render or the calligraphers who write the character. Thus, numerous images can represent the same character. Our proposed method aims to learn a way to generate Chinese character images with expected styles from a given character.
Let $h$ be a character code encoded by a system such as Unicode, $s$ be a style label, and $y$ be an image representing $h$ under the style $s$.
From $h$, we render an image $x$ through a given Chinese font. Thus, $x$'s style is assigned. We use the paired image sets $\{x\}$ and $\{y\}$ to train our networks to translate font-rendered images to calligrapher-written ones. 

\textbf{\flushleft Architecture.}
Figure~\ref{fig:model} shows the architecture of the proposed method. Given $h$, we render an image $x$ through a given Chinese font, and then we encode $x$ through an image encoder $E_i$ to generate an image feature vector $v_i$. At the same time, we consult a dictionary $T$ to obtain $h$'s component sequence $c$ to generate a component feature vector $v_c$ through a component encoder $E_c$. We convert the style label $s$ of the reference image $y$ to a one-hot vector $v_s$.
We concatenate $v_c$, $v_i$, and $v_s$ as an input feature vector used by an image decoder $G$ to generate a calligraphy character image $\hat{y}$.

To train $E_c$, $E_i$, and $G$, we use an addition networks an image pair discriminator $D$ and its auxiliary style classifier $D_s$. We explain their design and motivations as follows.

\textbf{\flushleft Image encoder and decoder.} We use U-Net~\cite{unet} as our encoder-decoder architecture, in a way similar to two existing image translation methods---pix2pix and zi2zi~\cite{pix2pix,zi2zi}.
Because Chinese calligraphy is mostly performed in black ink, we assume our images are grayscale without colors. Thus, we slightly modify U-Net's architecture by reducing the channel number of the input and output images from 3 to 1. Because our image decoder $G$ requires $v_s$ and $v_c$ as additional input data, we lengthen the length of $G$'s input vector. Table~\ref{table:architecture_g} shows the proposed architecture.
\begin{table}[t]
\centering
\begin{tabular}{ccc}
& \multicolumn{2}{c}{Shape}\\
\hline
Layer & Encoder & Decoder\\
\hline
Input & \(256 \times 256 \times 1\) & \(1 \times 1 \times 775\)\\
L1 & \(128 \times 128 \times 64\) & \(2 \times 2 \times 512\)\\
L2 & \(64 \times 64 \times 128\) & \(4 \times 4 \times 512\)\\
L3 & \(32 \times 32 \times 256\) & \(8 \times 8 \times 512\)\\
L4 & \(16 \times 16 \times 512\) & \(16 \times 16 \times 512\)\\
L5 & \(8 \times 8 \times 512\) & \(32 \times 32 \times 256\)\\
L6 & \(4 \times 4 \times 512\) & \(64 \times 64 \times 128\)\\
L7 & \(2 \times 2 \times 512\) & \(128 \times 128 \times 64\)\\
L8 & \(1 \times 1 \times 512\) & \(256 \times 256 \times 1\)\\
\hline
\end{tabular}
\caption{Architecture of the image encoder and decoder. All 8 encoder layers use the same convolution kernel size \(5 \times 5\), activation function LeakyReLU with a slope as 0.2, batch normalization layer, and stride size of 2. The decoder's L1 to L7 layers use the same deconvolution kernel size \(5 \times 5\), activation function ReLU, batch normalization layer. The decoder's L8 layer uses the hyperbolic tangent activation function and has a drop out layer with a drop rate as 0.5.}
\label{table:architecture_g}
\end{table}

\begin{figure}[t]
\centering
\includegraphics[width=0.6\linewidth]{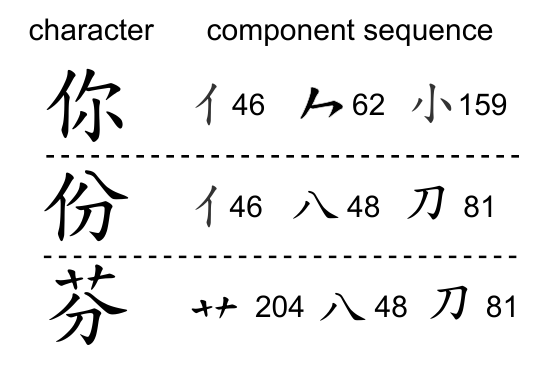}
\caption{Examples of component sequences. The first and second characters share the same component code $k_1$ as 46, and the second and third characters share the same $k_2$ as 48 and $k_3$ as 81.}
\label{figure:component_sequence}
\end{figure}
\begin{figure}[t]
\includegraphics[width=\linewidth]{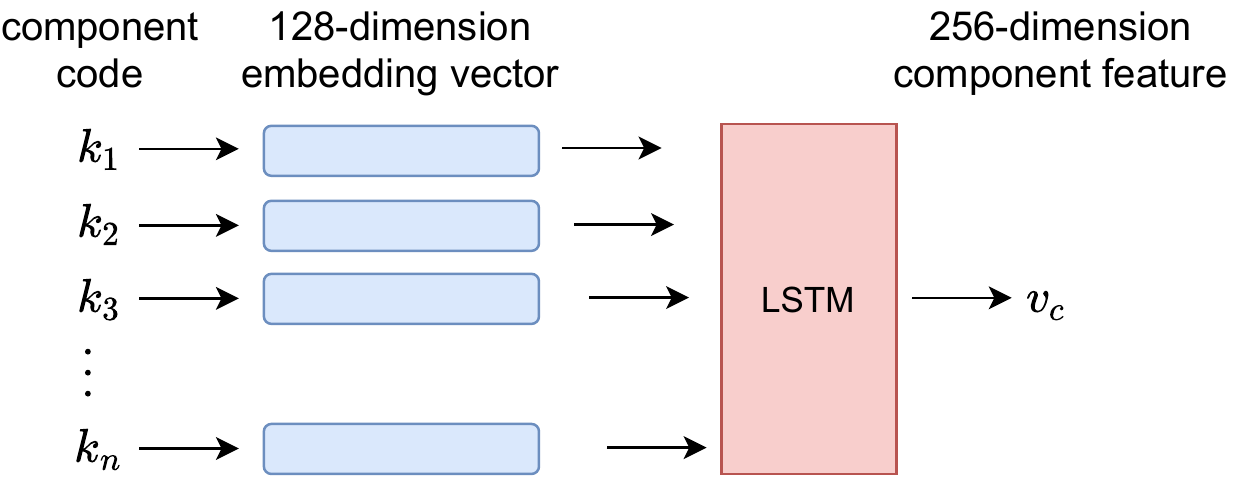}
\caption{Architecture of the proposed component encoder.}
\label{figure:component_encoder}
\end{figure}

{\bf \flushleft Component encoder.}
Chinese characters are composed of basic stroke and dot units. Their relative positions and intersections form numerous components, each consists of a few strokes and dots in specific shapes. That is the reason that Chinese characters are highly structured and the property that we exploit to develop our method. Figure~\ref{figure:component_sequence} shows a few examples of components.
We use a publicly available Chinese character decomposition system, the Chinese Standard Interchange Code\footnote{\url{https://www.cns11643.gov.tw}}, which defines
517 components most Chinese characters.
Given a character $h$, we use the system to obtain its component sequence 
\begin{equation}
    c = (k_1, k_2, ..., k_n),
\end{equation}
where $n$ is length of $c$ depending on $h$.
To convert the variable-length sequence $c$ to a fixed-length feature vector $v_c$, we propose a sequence encoder as shown in Figure~\ref{figure:component_encoder}, which contains an embedding layer and a LSTM model. The embedding layer converts component codes  to 128-dimension embedding vectors, which will be input to the LSTM model to generate a structure feature vector \(v_k\). Those embedding vectors are automatically optimized during our training process. We initialize the LSTM model randomly.

{\bf\flushleft Discriminator and auxiliary style classifier.}
Our discriminator and auxiliary style classifier are almost the same as the one used in zi2zi, except the channel number of the input layer. Its architecture is shown in Table~\ref{table:architecture_D}. The discriminator and auxiliary style classifier share the first three layers, and own independent 4th layers.
\begin{table}[t]
\begin{tabular}{ccc}
\hline
Layer & Type & Shape \\
\hline
Input & Image pair & \(256 \times 256 \times 2\) \\
L1 & Conv\(5 \times 5\), ReLU, BN & \(256 \times 256 \times 64\) \\
L2 & Conv\(5 \times 5\), ReLU, BN & \(128 \times 128 \times 128\) \\
L3 & Conv\(5 \times 5\), ReLU, BN  & \(64 \times 64 \times 256\) \\
L4-1 ($D$) & Linear & 1 \\
L4-2 ($D_s$) & Linear & 7 \\
\hline
\end{tabular}
\caption{Architecture of the proposed discriminator $D$ and style classifier $D_s$. BN means a batch normalization layer.}
\label{table:architecture_D}
\end{table}

\textbf{\flushleft Losses.} We define 4 losses to train our model. The adversarial loss of a conditional GAN
\begin{equation}
\mathcal{L}_{cGAN}=\log D(x, y) + \log (1-D(x, \hat{y}))
\label{eq:adv_loss}
\end{equation}
is use to help our generated images look realistic.
To encourage the generated images to be similar to the training ones, we use a pixel-wise
\begin{equation}
\label{eq:l1_loss}
\mathcal{L}_{p} = \|y-\hat{y}\|_1.
\end{equation}
Because the input image $x$ and output $\hat{y}$ represent the same character, we use a constancy loss in the same way as~ \cite{dtn,zi2zi}
\begin{equation}
\label{eq:constant_loss}
\mathcal{L}_c = \|E_i(x)-E_i(\hat{y})\|_1,
\end{equation}
which encourages the two images to have similar feature vectors. 
The generated images should retain the assign style, so we define a category loss
\begin{equation}
\label{eq:cls_loss}
\mathcal{L}_s = \log(D_s(s|y)) + \log(D_s(s|\hat{y})).
\end{equation} 
We set our full objective function
\begin{equation}
\mathcal{L} = \mathcal{L}_{cGAN} + \lambda_{p}\mathcal{L}_{p} + \lambda_c\mathcal{L}_c + \lambda_s\mathcal{L}_s
\end{equation}
where $\lambda_p$, $\lambda_c$, and $\lambda_s$ are parameters to control the relative importance of each loss.

\section{Experimental Validation}
To compile an image set to conduct experiments, we download images from a Chinese calligraphy character image repository\footnote{\url{http://163.20.160.14/~word/modules/myalbum/}}. All images are brush-written by an expert emulating ancient masterpieces, or rendered from art fonts.
The repository covers 29 calligraphy styles.
Some of them belong to the well-defined regular script, semi-cursive script, and clerical script by their names, but the remaining are not categorized.
We select the 7 styles belonging to regular script to conduct our experiments, and their names are
\begin{enumerate}[topsep=0pt,itemsep=-1ex,partopsep=1ex,parsep=1ex]
	\item Chu Suiliang,
	\item Liu Gongquan,
	\item Ouyang Xun--Huangfu Dan Stele,
	\item Ouyang Xun--Inscription on Sweet Wine Spring at 
Jiucheng Palace,
	\item Yan Zhenqing--Stele of the Abundant Treasure Pagoda,
	\item Yan Zhenqing--Yan Qinli Stele,
	\item Yu Shinan.
\end{enumerate}
\begin{figure}[t]
    \centering
    \begin{tabular}{@{}c@{}c@{}c@{}c@{}c@{}c@{}c@{}}
    \includegraphics[height=1.2cm]{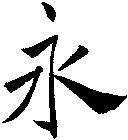} & 
    \includegraphics[width=1.2cm]{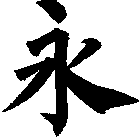} & 
    \includegraphics[height=1.2cm]{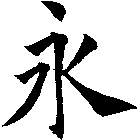} & 
    \includegraphics[width=1.2cm]{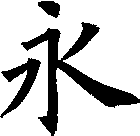} & 
    \includegraphics[width=1.2cm]{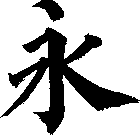} & 
    \includegraphics[height=1.2cm]{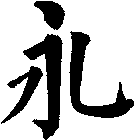} & 
    \includegraphics[height=1.2cm]{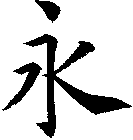} \\
    1 & 2 & 3 & 4 & 5 & 6 & 7 \\
    \end{tabular}
    \caption{Example characters of the 7 styles downloaded from the online repository. The 1st, 3rd, 6th, and 7th images have a vertical long side, while the 2nd, 4th, and 5th ones have a horizontal long side.}
    \label{fig:7_style_examples}
\end{figure}

The 3rd and 4th styles are created by the same ancient master calligrapher Ouyang Xun in his early and late years. Because of the change of the calligrapher's style, we treat them as two different styles, which is also the rule of thumb 
in the Chinese calligraphy community. The 5th and 6th styles are of the same case. Figure~\ref{fig:7_style_examples} shows examples of the 7 styles. 
There are several thousand images available for each style, but some images under the same style may represent the same character. In total, we collect 47552 images which covers 6548 different characters but only 5560 characters are available in all the 7 styles. Table~\ref{tab:Sample_Statistics} shows their statistics.
We randomly select 1000 characters out of the 5560 common characters set as our test character set, and have 7737 test images. We use the remaining 39815 images to train our model.
\begin{table}[t]
    \centering
    \begin{tabular}{@{\hspace{3pt}}c@{\hspace{3pt}}|@{\hspace{3pt}}c@{\hspace{3pt}}c@{\hspace{3pt}}c@{\hspace{3pt}}c@{\hspace{3pt}}c@{\hspace{3pt}}c@{\hspace{3pt}}c@{\hspace{3pt}}|@{\hspace{3pt}}c@{\hspace{3pt}}}
    \hline
        Style & 1 & 2 & 3 & 4 & 5 & 6 & 7 & Total\\
        \hline
        Training & 5975 & 5127 & 5873 & 5809 & 5283 & 5884 & 5864 & 39815 \\
        Test & 1184 & 1044 & 1126 & 1092 & 1025 & 1122 & 1144 & 7737 \\
        \hline
        Total & 7159 & 6171 & 6999 & 6901 & 6308 & 7006 & 7008 & 47552 \\
        \hline
    \end{tabular}
    \caption{Statistics of our training and test samples.}
    \label{tab:Sample_Statistics}
\end{table}

The repository's image size varies depending on character's shapes, but the long side is fixed 140 pixels. We keep its aspect ratio and enlarge the long side to 256 pixels through Lanczos resampling. We place the enlarged image at the center and pad the two short sides to generate a square image of 256$\times$256 pixels as our ground truth image $y$. The repository's image color depth is 1-bit monochrome. We do not change the depth during resampling. Our network linearly converts those monochrome images to tensors with a value range between -1 and 1. We use the font Sim Sun to render input images $x$ because it covers a large number of characters and it is used by zi2zi. Its rendered images are grayscale and show characters at the image center.

{\bf \flushleft Training setup.} 
We randomly initialize our networks' weights. We use the Adam~\cite{adam} optimizer to train our model with parameters $\beta_1$ as 0.5, $\beta_2$ as 0.999, and batch size 16. Because our discriminator $D$ learns faster than the generator does, we update the generator twice after updating the discriminator once.
We train our model in 40 epochs. We set the initial learning rate as 0.001 for the first 20 epochs and the decay rate as 0.5 for the following 20 epochs. It takes 25 hours to train our model on a machine equipped with an 8-core 2.1GHz CPU and an Nvidia GPU RTX 2080 Ti. We set $\lambda_p$ as 100, $\lambda_c$ as 15 and $\lambda_s$ as 1. We implement the proposed method using TensorFlow.

{\bf \flushleft Evaluation.}
We evaluate our generated images quantitatively and qualitatively. We use the mean square error (MSE) and structural similarity index (SSIM)~\cite{Wang04_ssim} to measure the similarity between ground truth and generated images. 
We conduct a survey of calligraphy experts and college students about our generated images. 

\begin{table}
\centering
\begin{tabular}{lcc}
\hline
Method & MSE & SSIM \\
\hline
zi2zi & 26.02 & 0.5781 \\
zi2zi + one-hot & 23.44 & 0.5969 \\
zi2zi + $E_c$ & 21.46 & 0.6101 \\
Proposed & \textbf{19.49} & \textbf{0.6147} \\
\hline
\end{tabular}
\caption{Performance comparison. One-hot means that we replace zi2zi's label embedding vector with our proposed simple one-hot vector. The symbol $E_c$ means the proposed component encoder. The proposed method equals to zi2zi (single channel) + one-hot + $E_c$}
\label{table:result}
\end{table}

\begin{figure}[t]
\begin{center}
\includegraphics[width=\linewidth]{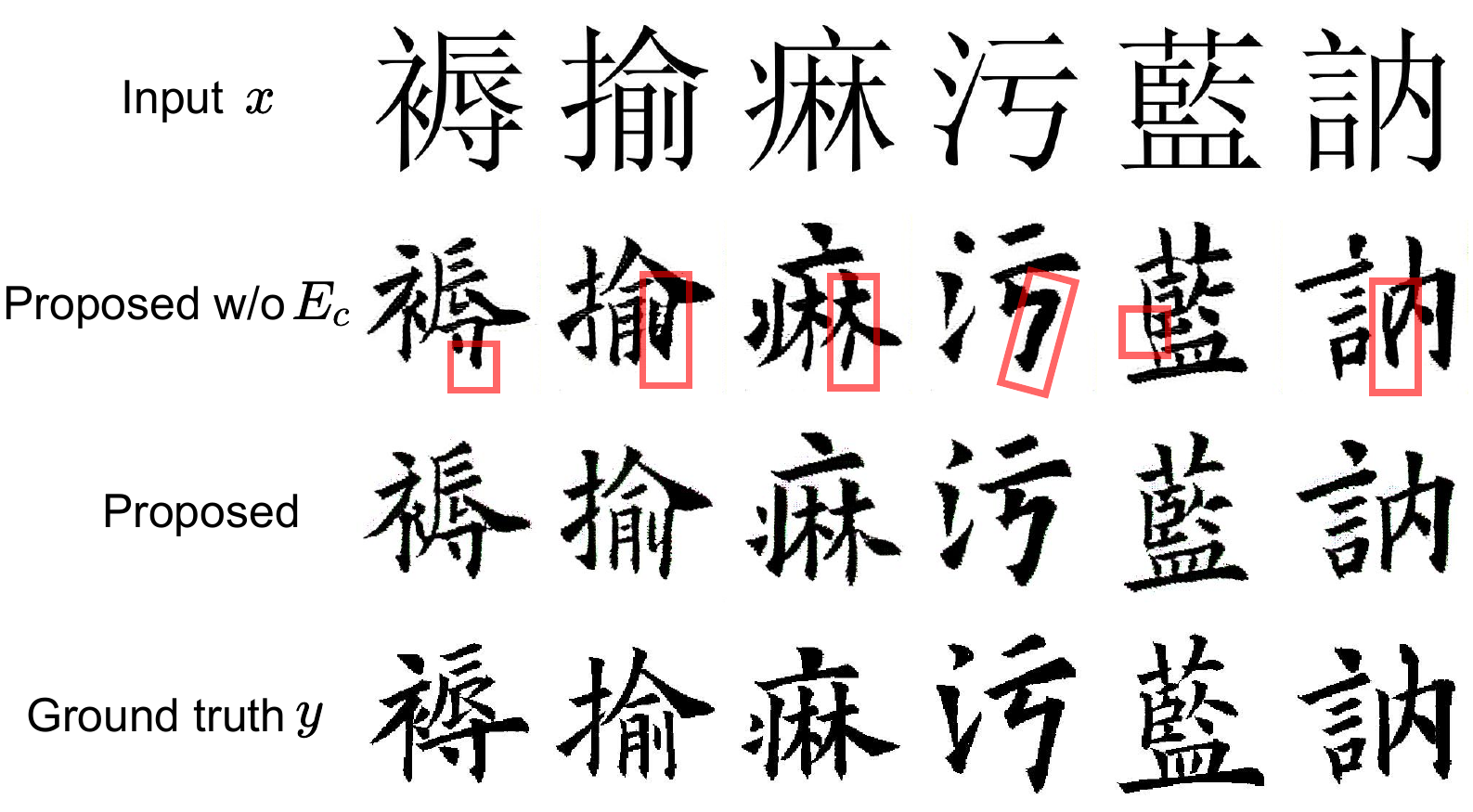}
\end{center}
   \caption{Qualitative comparison of single style transfer. All of the 6 characters are generated under the style 4. Red rectangles highlight the benefits brought by the proposed component encoder, which generates the ending hook of the first character, separates the two strokes of the second character, makes the strokes of the third, fourth, and sixth characters straight, and restores the corner of the L-shape stroke of the fifth character.}
\label{fig:effect_component_encoder}
\end{figure}

Table~\ref{table:result} shows the numerical evaluation of the proposed method, two weakened configurations, and a state-of-the-art method. The two weakened configurations are the two major differences between the proposed method and zi2zi, and the comparisons show that both of the proposed style and component encoders improve the generated images. Figure~\ref{fig:effect_component_encoder} shows a few examples of their qualitative differences. 

To further validate the proposed component encoder, we conduct another experiment of single style transfer. We remove the style feature $v_s$ and style loss from the proposed method to train 7 independent models and report their overall MSE and SSIM index in Table~\ref{table:result_no_style_labels}.

\begin{table}
\centering
\begin{tabular}{c|cc|cc}
\hline
 & \multicolumn{2}{c|}{MSE} & \multicolumn{2}{c}{SSIM} \\
Style & zi2zi & Proposed & zi2zi & Proposed\\
\hline
1 & 18.40 & \textbf{17.78} & 0.6230 & \textbf{0.6507} \\
2 & 19.46 & \textbf{18.13} & 0.6203 & \textbf{0.6513} \\
3 & 19.41 & \textbf{18.12} & 0.6446 & \textbf{0.6635} \\
4 & \textbf{19.02} & 20.15 & 0.6376 & \textbf{0.6485} \\
5 & \textbf{18.54} & 19.06 & 0.6382 & \textbf{0.6489} \\
6 & 17.76 & \textbf{17.57} & 0.6549 & \textbf{0.6628} \\
7 & \textbf{18.92} & 19.09 & 0.6179 & \textbf{0.6299} \\
\hline
mean & 18.79 & \textbf{18.56} & 0.6338 & \textbf{0.6508} \\
\hline
\end{tabular}
\caption{Quantitative comparison of single style transfer. We disable the multi-style part of both methods so the only considerable difference between the two configurations is the existence of a component encoder, which is contained in the proposed method, but not in zi2zi. For each style, training and test images used by the two methods are the same.}
\label{table:result_no_style_labels}
\end{table}

\begin{figure}[t]
\centering
\begin{tabular}{cccc}
$x$ & $y$ & zi2zi & Proposed\\
\includegraphics[width=0.2\linewidth]{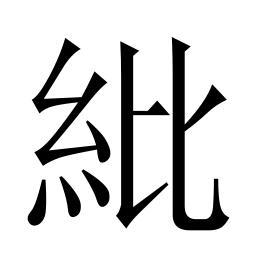}& \includegraphics[width=0.2\linewidth]{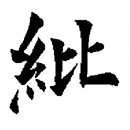} & \includegraphics[width=0.2\linewidth]{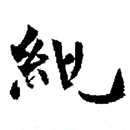} & \includegraphics[width=0.2\linewidth]{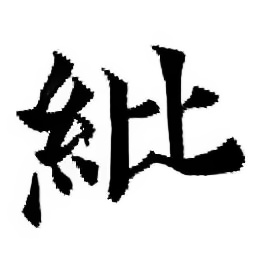} \\
\includegraphics[width=0.2\linewidth]{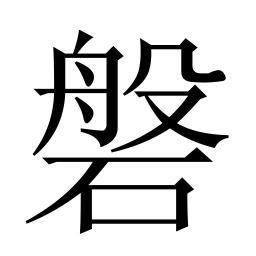} & \includegraphics[width=0.2\linewidth]{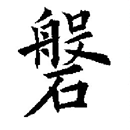} &
\includegraphics[width=0.2\linewidth]{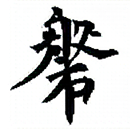} & \includegraphics[width=0.2\linewidth]{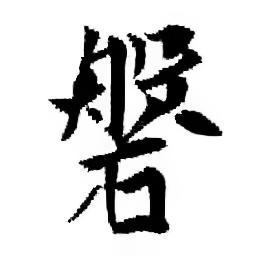} \\
\end{tabular}
   \caption{Failure cases generated by ziz2.}
\label{fig:broken_sample}
\end{figure}

For some difficult characters, we observe that zi2zi may generate failure images, as shown in Figure~\ref{fig:broken_sample}. That is the reason of its poor MSE and SSIM index. In contrast, the proposed method rarely generates failure images, and we attribute this improvement to the proposed component encoder.

\textbf{\flushleft Human subject study.}
Our human subjects are 18 undergraduate and graduate students, including 7 males and 11 females. All of them are Taiwanese, reading traditional Chinese characters every day. Among the 18 participants, 3 of them are members of a Chinese calligraphy club, 4 are not club members but have calligraphy skills learned in art classes, and 11 never use brushes to write Chinese characters. One participant is at the age of 40, and all others are between 20 and 30. For each participant, we randomly select 2 characters out of our 1000 test characters to generate images using both zi2zi and the proposed method. Because zi2zi may generate failure images, we intentionally skip that case. We generate images under all 7 styles so a participant sees 30 images, including 14 generated by zi2zi, 14 generated by the proposed method, and 2 of ground truth. We ask participants' opinions which image is more similar to the ground truth one. Table~\ref{table:perceptual_survey} shows the study's results. 

\begin{table}[t]
    \centering
    \begin{tabular}{c@{\hspace{7pt}}c@{\hspace{7pt}}c@{\hspace{7pt}}c@{\hspace{7pt}}c@{\hspace{7pt}}c@{\hspace{7pt}}c@{\hspace{7pt}}c}
    \hline
        Style & 1 & 2 & 3 & 4 & 5 & 6 & 7\\
        \hline
        zi2zi & 19.5 & 19.5 & 19.5 & 2.8 & 11.2 & 11.2 & 8.3 \\
        Proposed & \textbf{80.5} & \textbf{80.5} & \textbf{80.5} & \textbf{97.2} & \textbf{88.8} & \textbf{88.8} & \textbf{91.7} \\
        \hline
    \end{tabular}
    \caption{Percentage of preferred images of our human subject study. Most of our participants think the proposed method's output images are more similar to the ground truth than zi2zi's ones are.}
    \label{table:perceptual_survey}
\end{table}

\begin{figure}[t]
\includegraphics[width=1\linewidth]{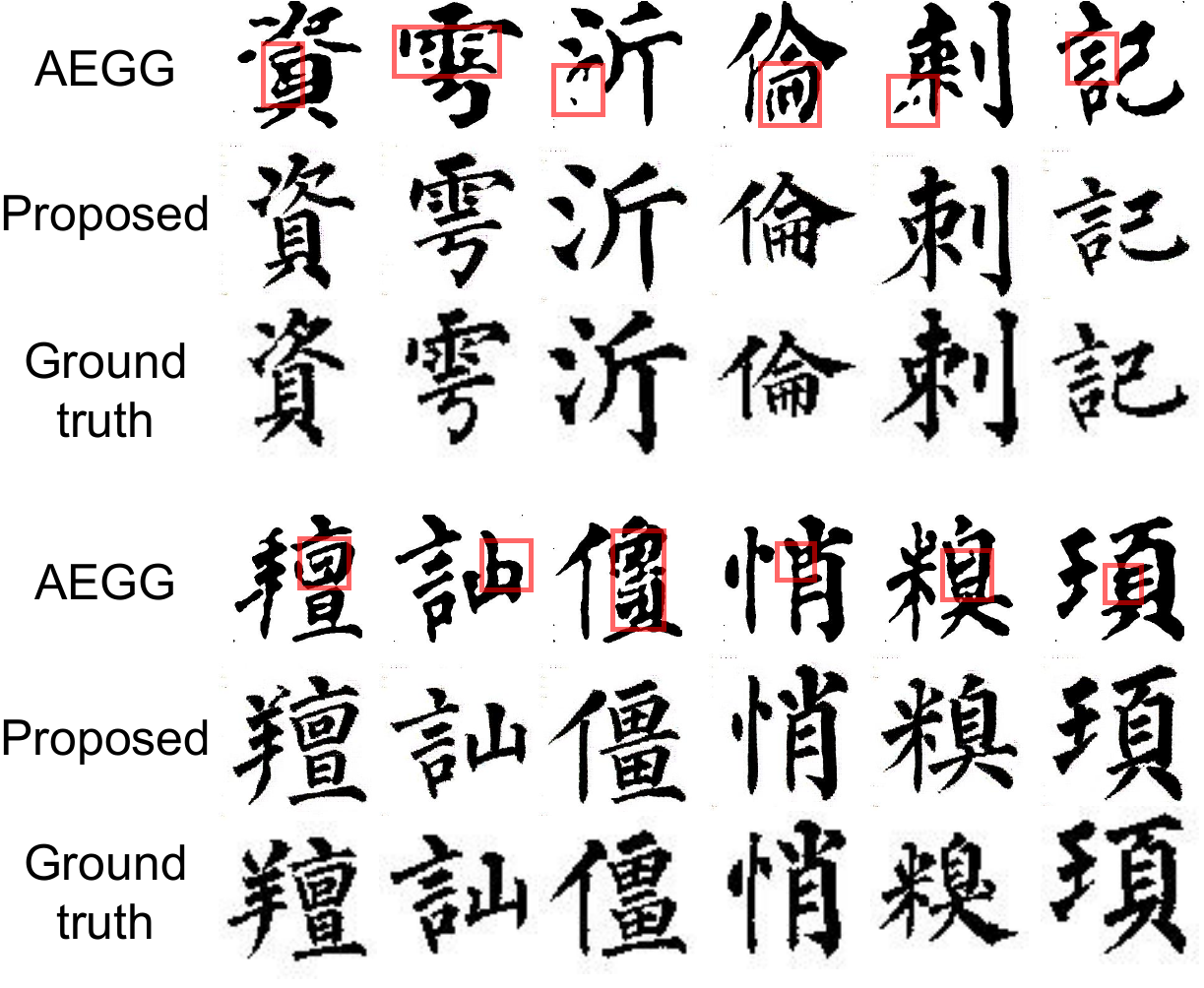}
\caption{Comparison with AEGG. The style used in this comparison is style 2. Those images generated by AEGG are extracted from its original paper. Their aspect ratios differ from the one of the ground truth images because AEGG's authors change the ratios. However, they do not explain the reason in their paper. Red rectangles highlight the regions that the proposed method handles better.}
\label{fig:compare_aegg}
\end{figure}

\textbf{\flushleft Comparison with AEGG.}
AEGG uses the same image repository as ours, and it is the only existing method to the best of our knowledge doing experiments using calligrapher-written images rather than font-rendered images, but its code and dataset are not publicly available.
Because we cannot get AEGG's dataset, we are unable to conduct a head-to-head comparison. However, the style used by AEGG is clearly specified in its paper, so we can still present rough comparisons to observe the general differences.
Because AEGG is a single-style transfer algorithm, we disable our multi-style part for a fair comparison.
Their results are shown in Figure~\ref{fig:compare_aegg}. The images generated by the proposed method show better structures (clearer intersections and less broken strokes) and richer details than the ones generated by AEGG.

\section{Conclusion and Future Study}
In this paper, we propose a novel method to generate multi-style Chinese character images. It consists a U-Net-based generator and a component encoder. Experimental results show that the proposed method generates high-quality images of calligraphy characters. Numerical evaluations and a human subject study show that the images generated by the proposed method more effectively than existing methods generates images similar to the ground truth ones.

Our research is still ongoing and many questions are not yet answered. For example, how well does the proposed method perform using other types of character images such as font-rendered images or images of cursive or semi-cursive scripts? Is there a font better than Sim Sun to render our input images? Does the choice depend on the used calligraphy styles? How many dimensions should we use for the component codes' embedding? Is there any pattern of those embedded feature vectors? Can some GAN training method such as WGAN-GP~\cite{wgan-gp} or SN-GAN~\cite{spectral_norm} improve our results? What is our method's performance if we use another data split? If we replace our shallow discriminator with a powerful and deep pre-trained image classifier, can we get better results? We wish we will be able to answer those questions soon.

{\small
\bibliographystyle{ieee_fullname}
\bibliography{egbib}

\begin{thebibliography}{10}\itemsep=-1pt

\bibitem{chang2018generating}
Bo Chang, Qiong Zhang, Shenyi Pan, and Lili Meng.
\newblock Generating handwritten {C}hinese characters using {CycleGAN}.
\newblock In {\em WACV}, 2018.

\bibitem{chang2018chinese}
Jie Chang, Yujun Gu, Ya Zhang, and Yan-Feng Wang.
\newblock Chinese handwriting imitation with hierarchical generative
  adversarial network.
\newblock In {\em BMVC}, 2018.

\bibitem{stargan}
Yunjey Choi, Minje Choi, Munyoung Kim, Jung-Woo Ha, Sunghun Kim, and Jaegul
  Choo.
\newblock {StarGAN}: Unified generative adversarial networks for multi-domain
  image-to-image translation.
\newblock In {\em CVPR}, 2018.

\bibitem{starganv2}
Yunjey Choi, Youngjung Uh, Jaejun Yoo, and Jung-Woo Ha.
\newblock {StarGAN} v2: Diverse image synthesis for multiple domains.
\newblock In {\em CVPR}, 2020.

\bibitem{dumoulin2016learned}
Vincent Dumoulin, Jonathon Shlens, and Manjunath Kudlur.
\newblock A learned representation for artistic style.
\newblock {\em arXiv preprint arXiv:1610.07629}, 2016.

\bibitem{gatys2016image}
Leon~A Gatys, Alexander~S Ecker, and Matthias Bethge.
\newblock Image style transfer using convolutional neural networks.
\newblock In {\em CVPR}, 2016.

\bibitem{wgan-gp}
Ishaan Gulrajani, Faruk Ahmed, Martin Arjovsky, Vincent Dumoulin, and Aaron~C
  Courville.
\newblock Improved training of {W}asserstein {GANs}.
\newblock In {\em NeurIPS}, 2017.

\bibitem{hinton2006reducing}
Geoffrey~E Hinton and Ruslan~R Salakhutdinov.
\newblock Reducing the dimensionality of data with neural networks.
\newblock {\em Science}, 313(5786):504--507, 2006.

\bibitem{huang2017arbitrary}
Xun Huang and Serge Belongie.
\newblock Arbitrary style transfer in real-time with adaptive instance
  normalization.
\newblock In {\em ICCV}, 2017.

\bibitem{pix2pix}
Phillip Isola, Jun-Yan Zhu, Tinghui Zhou, and Alexei~A Efros.
\newblock Image-to-image translation with conditional adversarial networks.
\newblock In {\em CVPR}, 2017.

\bibitem{jiang2017dcfont}
Yue Jiang, Zhouhui Lian, Yingmin Tang, and Jianguo Xiao.
\newblock {DCF}ont: an end-to-end deep {C}hinese font generation system.
\newblock In {\em SIGGRAPH Asia}. 2017.

\bibitem{jiang2019scfont}
Yue Jiang, Zhouhui Lian, Yingmin Tang, and Jianguo Xiao.
\newblock {SCFont}: Structure-guided {C}hinese font generation via deep stacked
  networks.
\newblock In {\em AAAI}, 2019.

\bibitem{johnson2016perceptual}
Justin Johnson, Alexandre Alahi, and Li Fei-Fei.
\newblock Perceptual losses for real-time style transfer and super-resolution.
\newblock In {\em ECCV}, 2016.

\bibitem{kim2017learning}
Taeksoo Kim, Moonsu Cha, Hyunsoo Kim, Jung~Kwon Lee, and Jiwon Kim.
\newblock Learning to discover cross-domain relations with generative
  adversarial networks.
\newblock In {\em ICML}, 2017.

\bibitem{adam}
Diederik~P Kingma and Jimmy Ba.
\newblock Adam: A method for stochastic optimization.
\newblock {\em arXiv preprint arXiv:1412.6980}, 2014.

\bibitem{Lian2016}
Zhouhui Lian, Bo Zha, and Jianguo Xiao.
\newblock Automatic generation of large-scale handwriting fonts via style
  learning.
\newblock In {\em SIGGRAPH Asia}, 2016.

\bibitem{liu2017unsupervised}
Ming-Yu Liu, Thomas Breuel, and Jan Kautz.
\newblock Unsupervised image-to-image translation networks.
\newblock In {\em NeurIPS}, 2017.

\bibitem{aegg}
Pengyuan Lyu, Xiang Bai, Cong Yao, Zhen Zhu, Tengteng Huang, and Wenyu Liu.
\newblock Auto-encoder guided {GAN} for {C}hinese calligraphy synthesis.
\newblock In {\em ICDAR}, 2017.

\bibitem{spectral_norm}
Takeru Miyato, Toshiki Kataoka, Masanori Koyama, and Yuichi Yoshida.
\newblock Spectral normalization for generative adversarial networks.
\newblock In {\em ICLR}, 2018.

\bibitem{acgan}
Augustus Odena, Christopher Olah, and Jonathon Shlens.
\newblock Conditional image synthesis with auxiliary classifier {GAN}s.
\newblock In {\em ICML}, 2017.

\bibitem{unet}
Olaf Ronneberger, Philipp Fischer, and Thomas Brox.
\newblock U-{N}et: Convolutional networks for biomedical image segmentation.
\newblock In {\em MICCAI}, 2015.

\bibitem{sun2017learning}
Danyang Sun, Tongzheng Ren, Chongxun Li, Hang Su, and Jun Zhu.
\newblock Learning to write stylized {C}hinese characters by reading a handful
  of examples.
\newblock In {\em IJCAI}, 2018.

\bibitem{sun2018pyramid}
Donghui Sun, Qing Zhang, and Jun Yang.
\newblock Pyramid embedded generative adversarial network for automated font
  generation.
\newblock In {\em ICPR}, 2018.

\bibitem{dtn}
Yaniv Taigman, Adam Polyak, and Lior Wolf.
\newblock Unsupervised cross-domain image generation.
\newblock {\em arXiv preprint arXiv:1611.02200}, 2016.

\bibitem{zi2zi}
Yuchen Tian.
\newblock zi2zi: Master {C}hinese calligraphy with conditional adversarial
  networks.
\newblock \url{https://kaonashi-tyc.github.io/2017/04/06/zi2zi.html}, 2017.

\bibitem{ulyanov2016texture}
Dmitry Ulyanov, Vadim Lebedev, Andrea Vedaldi, and Victor~S Lempitsky.
\newblock Texture networks: Feed-forward synthesis of textures and stylized
  images.
\newblock In {\em ICML}, 2016.

\bibitem{Wang04_ssim}
Zhou Wang, Alan~C. Bovik, Hamid~R. Sheikh, and Eero~P. Simoncelli.
\newblock Image quality assessment: from error visibility to structural
  similarity.
\newblock {\em TIP}, 13(4):600--612, 2004.

\bibitem{wen2019handwritten}
Chuan Wen, Jie Chang, and Ya Zhang.
\newblock Handwritten {C}hinese font generation with collaborative stroke
  refinement.
\newblock {\em arXiv preprint arXiv:1904.13268}, 2019.

\bibitem{xu2009automatic}
Songhua Xu, Hao Jiang, Tao Jin, Francis~CM Lau, and Yunhe Pan.
\newblock Automatic generation of {C}hinese calligraphic writings with style
  imitation.
\newblock {\em IEEE Intelligent Systems}, (2):44--53, 2009.

\bibitem{xu2007intelligent}
Songhua Xu, Hao Jiang, Francis Chi-Moon Lau, and Yunhe Pan.
\newblock An intelligent system for {C}hinese calligraphy.
\newblock In {\em AAAI}, 2007.

\bibitem{xu2005automatic}
Songhua Xu, Francis~CM Lau, William~K Cheung, and Yunhe Pan.
\newblock Automatic generation of artistic {C}hinese calligraphy.
\newblock {\em IEEE Intelligent Systems}, 20(3):32--39, 2005.

\bibitem{yi2017dualgan}
Zili Yi, Hao Zhang, Ping Tan, and Minglun Gong.
\newblock {DualGAN}: Unsupervised dual learning for image-to-image translation.
\newblock In {\em ICCV}, 2017.

\bibitem{zhang2018separating}
Yexun Zhang, Ya Zhang, and Wenbin Cai.
\newblock Separating style and content for generalized style transfer.
\newblock In {\em CVPR}, 2018.

\bibitem{zhu2017unpaired}
Jun-Yan Zhu, Taesung Park, Phillip Isola, and Alexei~A Efros.
\newblock Unpaired image-to-image translation using cycle-consistent
  adversarial networks.
\newblock In {\em ICCV}, 2017.

\end{thebibliography}
}

\end{document}